\documentclass[letterpaper, 10 pt, conference]{ieeeconf}  % Comment this line out
\IEEEoverridecommandlockouts                              % This command is only
\usepackage{amsmath}
\usepackage{amssymb}
\usepackage{amsfonts}
\usepackage[pdftex]{graphicx}
\DeclareGraphicsExtensions{.pdf, .png, .jpg}
\usepackage{algorithm}
\usepackage[noend]{algpseudocode}
\algrenewcommand\algorithmicindent{0.7em}%
\usepackage{gensymb}
\usepackage[caption=false]{subfig}
\usepackage{cite}
\usepackage{mathtools,bm}

\usepackage{amsthm}
\newtheorem{problem}{Problem}

\newtheorem{definition}{Definition}

\DeclareMathOperator*{\argmax}{argmax}

\title{\LARGE \bf
Signal Temporal Logic Synthesis as Probabilistic Inference
}
\author{Ki Myung Brian Lee, Chanyeol Yoo and Robert Fitch % => this % stops a space
\thanks{This research is supported by an Australian Government Research Training Program (RTP) Scholarship and the University of Technology Sydney.}%
\thanks{Authors are with the University of Technology Sydney, Ultimo, NSW 2006, Australia {\tt\footnotesize brian.lee@student.uts.edu.au, \{chanyeol.yoo, rfitch\}@uts.edu.au}}
}

\begin{document}

\maketitle

\begin{abstract}
We reformulate the signal temporal logic~(STL) synthesis problem as a maximum a-posteriori~(MAP) inference problem.
To this end, we introduce the notion of random STL~(RSTL), which extends deterministic STL with random predicates.
This new probabilistic extension naturally leads to a synthesis-as-inference approach.
The proposed method allows for differentiable, gradient-based synthesis while extending the class of possible uncertain semantics.
We demonstrate that the proposed framework scales well with GPU-acceleration, and present realistic applications of uncertain semantics in robotics that involve target tracking and the use of occupancy grids.
\end{abstract}

%%%%%%%%%%%%%%%%%%%%%%%%%%%%%%%%%%%%%%%%%%%%%%%%%%
%%%%%%%%%%%%%%%%%%%%%%%%%%%%%%%%%%%%%%%%%%%%%%%%%%
\section{Introduction} 
% Vision
% STL synthesis will allow intuitive human-robot commanding interface
% We want to do TL in actual real life, which hasn't been done
Temporal logic is a promising tool for robotics applications and explainable AI in that it can be used to represent rich, complex task objectives in the form of human-readable logical specifications. Robot systems equipped with the capability to perform STL synthesis can implement powerful, intuitive command interfaces using STL. We are interested in developing STL synthesis for practical, real-world applications by viewing the synthesis problem as a form of probabilistic inference.

% Motivation
More widespread adoption of formal methods in robot systems, in our view, is limited by two main factors. First, the computational requirements of existing synthesis methods are seen as prohibitive. Second, achieving computational efficiency is seen to compromise semantic expressivity.

% % Challenge
% % Existing methods are computationally intensive 
% % Existing methods are limited in terms of semantic expressiveness and computation time. 
% Existing methods have two limitations
% Doesn't consider probabilistic
% Too slow

% Solution
% \begin{itemize}
%     \item Viewing STL synthesis as probabilistic inference solves both these problems at once.
%     \item Propose the notion of random STL
%     \item Propose three approximate methods for computing probability of random STL
%     \item Solve these with gradient ascent with GPU-acceleration
%     \item Present empirical analysis of correctness, convergence, and scalability.  
% \end{itemize}
The approach we adopt in this paper is to address challenges in both computational efficiency and expressivity by viewing STL synthesis as probabilistic inference. 
This probabilistic approach fits naturally with robotics perception-action pipelines for important information gathering tasks such as search, target tracking, and mapping.

We first present a probabilistic extension of STL, called random STL (RSTL), that is designed to support synthesis of robot trajectories that satisfy specifications defined over uncertain events in the environment.
Formally, RSTL extends STL to include \emph{uncertain} semantic labels.
The gain in expressivity is the capacity to specify tasks in a way that facilitates robust behaviour in real-world settings; uncertainty that is inherent to practical environments can be anticipated explicitly instead of handled reactively.  
Synthesising trajectories is a differentiable problem that largely resembles previously proposed differentiable measures of robustness for deterministic STL synthesis.

We then present three approximate methods for computing the probability of satisfaction of RSTL formulae. Incorporating these methods, we implement synthesis using GPU-accelerated gradient ascent and output the most probable sequence of control actions to satisfy the given specification. 

To evaluate our method, we report empirical results that illustrate correctness, convergence, and scalability properties. Further, our method inherits the benefits of gradient-based MPC, including the anytime property and predictable computation time per iteration.

To demonstrate practical use in common robotics scenarios, we provide case studies involving target search and occupancy grids. These examples show that desirable behaviour naturally arises, such as prioritising targets whose location uncertainty is increasing over targets that are physically proximate. They also demonstrate the use of complex predicates such occupancy grids which are prevalent robotics applications. Computational efficiency is shown to parallel recent advances in approximate synthesis for deterministic STL and, importantly, can be further improved through additional GPU hardware to enable development of highly capable robot systems.

We view the main contribution of this work as a step towards the feasibility of temporal logic for robotics in practice through the introduction of a new method for synthesis as probabilistic inference that can accommodate powerful task specifications and that has useful performance characteristics. Our work also provides the basis for interesting theoretical extensions that would allow temporal logic specifications to be integrated with estimation methods and belief-based planning.

\section{Related Work} 
To model tasks in uncertain environments, one approach is to model the robots' dynamics as a discrete Markov decision process~(MDP), where each state is assigned semantic labels with corresponding uncertainty~\cite{uncertain_ltl_1,takashi_tanaka,yoo2012probabilistic,yoo2013provably}.
% In this setting, linear temporal logic~(LTL)~\cite{} is the natural specification of tasks. 
A natural task specification tool is linear temporal logic~(LTL), given which a product MDP~\cite{uncertain_ltl_1} is constructed from an automaton and a sequence of actions are found that maximize the probability of satisfaction.  
These ideas can be extended to the continuous case by judiciously partitioning the environment~\cite{tulip,yoo2016online,lee2020hierarchical}. 
However, we find that partitioning is computationally prohibitive for online operations in uncertain environments, because any change in belief about the environment would lead to invalidation and expensive recomputation. 
Moreover, the construction of a product MDP is a computationally expensive operation, and improving its scalability remains an open problem. 

Signal temporal logic~(STL)~\cite{donze_stl} is defined over continuous signals. Unlike in LTL, satisfaction is determined using continuous-valued \emph{robustness}~\cite{donze_monitoring}. Existing work uses STL to specify a task defined over a set of deterministic classes of conditions on the environment uncertainty, such as chance constraints or variance limits~\cite{Sadigh-RSS-16,stream,belief_space}.
However, since the robustness evaluation is not differentiable, a common approach is to synthesise solutions using \emph{mixed integer linear programming} (MILP) which scales exponentially with the size of mission horizon~\cite{Sadigh-RSS-16,raman_mpc}. To address the inherent complexity, the robustness metric is approximated to smooth the search space and find a solution using gradient ascent.

\section{Problem Formulation}

Suppose we have a robot with $N$-dimensional state~$\mathbf{x}_{t} \in \mathbb{R}^N$ and  control actions $\mathbf{u}_{t} \in \mathbb{U}$, where $\mathbb{U}$ is a continuous set of admissible control actions. 
The robot's dynamics is uncertain, and is modelled by a discrete-time, continuous-space MDP~$\mathcal{P}(\mathbf{x}_{t+1} \mid \mathbf{x}_{t}, \mathbf{u}_{t})$ between $t$ and $t+1$, so that the trajectory distribution over a horizon $T$ is given by:
\begin{equation}\label{eq:dynamic_model}
    \mathcal{P}(\mathbf{X} \mid \mathbf{U} ) = \mathcal{P}(\mathbf{x}_{1}) \prod_{t=1}^{T} \mathcal{P}(\mathbf{x}_{t+1} \mid \mathbf{x}_{t}, \mathbf{u}_{t}),
\end{equation}
where $\mathbf{X}\equiv\mathbf{x}_{1}...\mathbf{x}_{T}$ and $\mathbf{U}\equiv\mathbf{u}_{1}...\mathbf{u}_{T}$.

The robot encounters a finite set of random events~$\mathcal{E}=\{E^{1}, ..., E^{M}\}$~(e.g., `object detected'),
whose probability of occurrence depends on robot's state $\mathbf{x}_{t}$ and time $t$.
We are interested in finding control actions~$\mathbf{U}^{*}$ that maximises the probability of satisfying a task~$\Phi$ defined over~$\mathcal{E}$~(e.g. `detect all objects').
Namely, this is a synthesis problem: 
\begin{problem}[Synthesis] \label{problem:synthesis}
    Given the uncertain dynamics~\eqref{eq:dynamic_model}, and a task specification~$\Phi$ over a set of random events~$\mathcal{E}$ with probability of satisfaction $\mathcal{P}\left( (\mathbf{X}, t) \models \Phi \right) $, find an optimal sequence of controls~$\mathbf{U}^*$ such that the trajectory $\mathbf{X}$ maximises the probability of satisfying~$\Phi$ over time horizon~$T$:
    \begin{equation}
        \mathbf{U}^{*} = \argmax_{\mathbf{U} \in \mathbb{U}^{T}}~\mathcal{P}\left( (\mathbf{X}, t) \models \Phi \right)    
        ,
    \end{equation}
    with respect to time~$t = 1$.
\end{problem}
% \noindent Admittedly, Problem~\ref{problem:synthesis} lacks a definition of task $\Phi$ and its satisfaction probability $\mathbb{P}\left(x, t, \Phi\right)$.
% We define these in Sec.~\ref{sec:meat:rstl}.

%%%%%%%%%%%%%%%%%%%%%%%%%%%%%%%%%%%%%%%%%%%%%%%%%%
%%%%%%%%%%%%%%%%%%%%%%%%%%%%%%%%%%%%%%%%%%%%%%%%%%
\section{STL Synthesis as Probabilistic Inference} 
% We propose a probabilistic task planning framework where we consider Problem~\ref{problem:pf} as probabilistic inference problem. 
In this section, we first introduce \emph{random signal temporal logic}~(RSTL), a probabilistic extension of STL that allows specification of tasks~$\Phi$ over random events~$\mathcal{E}$. 
We then present a probabilistic inference formulation of Problem~\ref{problem:synthesis}.
% We then present a  of RSTL that allows solving synthesis as inference. 
% Then, we discuss we could treat the synthesis problem as probabilistic inference.
% We propose  \emph{random STL formula}~(RSTL) to model temporal logic formulae over uncertain events.
% The probabilistic 
% This motivates solving the synthesis problem as probabilistic inference.
% \subsection{Random Signal Temporal Logic}\label{sec:meat:rstl}

%%%%%%%%%%%%%%%%%%%%%%%%%%%%%%%%%%%%%%%%%%%%%%%%%%
\subsection{Random STL Formulae}\label{sec:meat:rstl}
% To extend the notion of deterministic STL formulae to the probabilistic domain, we define \emph{random STL formulae} as follows. 
We model the random events $\mathcal{E}=\{E^{1}, ..., E^{M}\}$ as (not necessarily independent) Bernoulli random variables that are dependent on robot's state and time, with conditional probability of occurrence:
\begin{equation}\label{eq:cond_bern}
    \mathcal{P}(E^{i} = 1 \mid \mathbf{x}_{t}, t) = \mathcal{P}^{i}(\mathbf{x}_{t}, t).
\end{equation}
In other words, each $E^{i}$ is a Bernoulli \emph{random field} over $\mathbb{R}^{N} \times \mathbb{R}^{+}$.
Given a set of random events $\mathcal{E}$, the syntax of an RSTL formula~$\Phi$ is given by: 
\begin{equation}
    \Phi := E \mid \neg \Phi \mid \Phi \wedge \Psi \mid \Phi \mathcal{U}_{[t_{1}, t_{2}]} \Psi
    ,
\end{equation}
where $E \in \mathcal{E}$, and $\Psi, \Phi$ are RSTL formulae. 
$\neg$ is logical negation, $\wedge$ is logical conjunction. 
$\mathcal{U}$ is the temporal operator `Until', and $\Phi \mathcal{U}_{[t_1,t_2]} \Psi$ means $\Phi$ must hold true between time~$[t_1,t_2]$ until~$\Psi$. 
Other operators such as~$\vee$ (disjunction), $\mathcal{F}_{[t_1,t_2]}$ (`in \textbf{F}uture', i.e., eventually) and~$\mathcal{G}_{[t_1,t_2]}$ (`\textbf{G}lobally', i.e., always) can be derived from the syntax the same way as deterministic STL~\cite{donze_stl}.
The events $E$ will be referred to as `event predicates'. 

RSTL is \emph{random} in the sense that, for a given trajectory $\mathbf{X}$, the satisfaction of an RSTL formula~$\Phi$ is a Bernoulli random event.
The probability of satisfaction is the \emph{expected} rate of satisfaction computed over sampled instances of the event predicates $\hat{\mathcal{E}} = \{ e_{1}, ..., e_{M} \} \sim \mathcal{E}$:
\begin{equation}\label{eq:prob_sat_monte_carlo}
    % \mathcal{P}( (\mathbf{x}, t) \models \Phi ) = \underset{\phi \sim \Phi}{\mathbb{E}}\mathbf{1}[ (x, t) \models \phi ]
    \mathcal{P}( (\mathbf{X}, t) \models \Phi ) = \underset{\hat{\mathcal{E}} \sim \mathcal{E}}{\mathbb{E}}\textbf{Sat}(\mathbf{X}, t, \hat{\mathcal{E}}, \Phi)
    ,
\end{equation}
where $\textbf{Sat}(\mathbf{X}, t, \hat{\mathcal{E}}, \Phi)$ is a deterministic function evaluated recursively as:
\begin{equation} \begin{aligned}
    \textbf{Sat}(\mathbf{X}, t, \hat{\mathcal{E}}, E^{i})            &\equiv e^{i}(\mathbf{x}_{t}, t) \sim \mathcal{P}^{i}(\mathbf{x}_{t}, t)\\
    \textbf{Sat}(\mathbf{X}, t, \hat{\mathcal{E}}, \neg \Phi)        &\equiv \neg \textbf{Sat}(\mathbf{X}, t, \hat{\mathcal{E}}, \Phi) \\
    \textbf{Sat}(\mathbf{X}, t, \hat{\mathcal{E}}, \Phi \wedge \Psi) &\equiv \textbf{Sat}(\mathbf{X}, t, \hat{\mathcal{E}}, \Phi) \wedge \textbf{Sat}(\mathbf{X}, t, \hat{\mathcal{E}}, \Psi) \\
    % &\hspace{2ex} \\
    \textbf{Sat}(\mathbf{X}, t, \hat{\mathcal{E}}, \Phi \mathcal{U}_{[t_{1}, t_{2}]} \Psi) &\equiv\\
    \bigvee_{\tau_{1} \in t + [t_{1}, t_{2}]}  
    \bigwedge_{\tau_{2} \in t_{1} + [0, \tau_{1}]} &\textbf{Sat}(\mathbf{X}, \tau_{2}, \hat{\mathcal{E}}, \Phi) \vee \textbf{Sat}(\mathbf{X}, \tau_{1}, \hat{\mathcal{E}}, \Psi).
\end{aligned} \end{equation}
% $\mathbf{x}_{1:T} \models \Phi$ is a random event with probability computed over instances of $E$.
% over event predicates $E \in \mathcal{E}$
% of satisfaction in Problem~\ref{problem:pf} is defined as the expected satisfaction over Bernoulli random events, such that
% Given an RSTL formula~$\Phi$ over a event predicates~$\mathcal{E}$ and a sequence of robot states~$\mathbf{x}_{1:T}$, 
Note that $e^{i}(\mathbf{x}_{t}, t) \in \{0, 1\}$ is a sample from $E^{i}$ at robot state $\mathbf{x}_{t}$ and time $t$. 

% Given state sequence~$\mathbf{x}$, time~$t$ and a set of sampled event instances~$\hat{\mathcal{E}}$, the satisfaction of formula~$\Phi$ can be evaluated deterministically using conventional STL robustness~\cite{donze_stl}. 
%%%%%%%%%%%%%%%%%%%%%%%%%%%%%%%%%%%%%%%%%%%%%%%%%%
\subsection{STL Synthesis as Inference}
% We now formally state the STL synthesis as an inference problem. 
% Control-as-inference paradigm~\cite{kappen,levine} has been shown to not only solve existing optimal control problems, 
% but also extend 
Since both task satisfaction and robot dynamics are probabilistic, it is natural to ask if Problem~\ref{problem:synthesis} can be solved solely within the realm of probability theory. 
This is achieved by the control-as-inference paradigm~\cite{kappen,levine}, which has been shown to not only encompass existing optimal control problems, but also to enable new approaches.
We follow a similar development and present an inference formulation of Problem~\ref{problem:synthesis}.

In this formulation, the problem is modelled by the joint distribution among task satisfaction, robot trajectory, and control actions:
% interpretation of the probability of satisfaction~\eqref{eq:prob_sat_monte_carlo}:
\begin{equation}
    \mathcal{P}( \Phi_{t}, \mathbf{X}, \mathbf{U}) = \mathcal{P}( \Phi_{t} \mid \mathbf{X}) \mathcal{P}(\mathbf{X} \mid \mathbf{U}) \mathcal{P}(\mathbf{U} ),
\end{equation}
where $\mathcal{P}( \Phi_{t} \mid \mathbf{X} ) \equiv \mathcal{P}( (\mathbf{X}, t) \models \Phi)$ denotes the probability satisfaction of $\Phi$ given $\mathbf{X}$ with respect to time $t$. 

Here, $\mathcal{P}(\mathbf{U})$ is our prior belief on what the control actions should be, and is representative of the admissible control space $\mathbb{U}$ in the synthesis formulation. 
For example, if $\mathcal{P}(\mathbf{U})$ is a zero-mean Gaussian prior, it is equivalent to penalising quadratic control cost.
The prior can derive from other knowledge, e.g., an imitation-learnt prior as~\cite{rowan} does for optimal control. 

A balance between admissibility and probability of satisfaction is captured by the \emph{posterior} probability of control actions given that the task is satisfied:
\begin{equation}\begin{aligned}
    \mathcal{P}(\mathbf{U} \mid \Phi_{t}) &\propto \mathcal{P}(\Phi_{t} \mid \mathbf{U} ) \mathcal{P}(\mathbf{U} ) \\
                                 &= \mathcal{P}(\mathbf{U}) \int \mathcal{P}(\Phi_{t} \mid \mathbf{X}) \mathcal{P}(\mathbf{X} \mid \mathbf{U})d\mathbf{X},
\end{aligned}
\end{equation}
% which can be interpreted as a balance between 
% $\mathcal{P}(\Phi_{0} \mid u )$ is the probability that $\Phi$ is satisfied given the control actions $u$. 
% Now, we state the inference analogue of Problem~\ref{problem:synthesis}. 
We thus pose Problem~\ref{problem:synthesis} as a maximum a posteriori~(MAP) inference problem: 
\begin{problem}[Inference]
Given the dynamic model~\eqref{eq:dynamic_model} and an RSTL task specification~$\Phi$, find the MAP control actions $\mathbf{U}^{*}$ given the robot's trajectory satisfies $\Phi$ with respect to $t$:
\begin{equation}\label{eq:stl_map}
% \begin{aligned}
    \mathbf{U}^{*} = \argmax_{\mathbf{U}} \mathcal{P}(\mathbf{U} \mid \Phi_{t}).
        %   &= \argmin_{u} -\log \mathcal{P}( \Phi_{0} \mid u) - \log \mathcal{P}(u)
% \end{aligned}
\end{equation}
\end{problem}

% \begin{remark}
% and a prior $\mathcal{P}(u)$, 
% An intuition behind~\eqref{eq:stl_map} is that it is a balance between two terms:
% We note that the choice of $u$ is not arbitrary; it may be subject to practical constraints, and we would like to minimise costs if possible. 
% We adopt the \emph{planning as inference} perspective~\cite{kappen,levine}, where a `prior knowledge' $\mathcal{P}(u)$ encodes that costly control actions shall not be likely. 
% \end{remark}
%%%%%%%%%%%%%%%%%%%%%%%%%%%%%%%%%%%%%%%%%%%%%%%%%%
%%%%%%%%%%%%%%%%%%%%%%%%%%%%%%%%%%%%%%%%%%%%%%%%%%
\section{Approximate Gradient Ascent}
In this section, we first present approximate methods that allow analytical evaluation of~\eqref{eq:prob_sat_monte_carlo}.
We then present a gradient-ascent scheme on these approximate evaluations. 
% In order to solve for~\eqref{eq:stl_map}, we present a framework where we approximate the evaluation function for tasks so that it becomes differentiable. 
% We then present a gradient-ascent scheme to solve for in~\eqref{eq:stl_map}.
% In this section, we present three approximate analytical methods for computing the probability of satisfaction~\eqref{eq:prob_sat_monte_carlo}.
% All of these methods are differentiable, and we use a gradient ascent scheme to solve the MAP problem~\eqref{eq:stl_map}.
\subsection{Conditional Independence Approximation}
To compute~\eqref{eq:prob_sat_monte_carlo} analytically, we observe that~\eqref{eq:prob_sat_monte_carlo} applies logical operations to samples from Bernoulli random variables. 
A convenient approximation is the product relation for independent Bernoulli random variables~$A$ and~$B$:
\begin{equation}\label{eq:independence}
    \mathcal{P}(A \wedge B) = \mathcal{P}(A)\mathcal{P}(B).
\end{equation}

Technically, the product relation holds true if the operands are \emph{conditionally independent} given $\mathbf{X}$. 
The conditionally independent~(CI)-approximation is defined by one of the authors~\cite{chanyeol_probabilistic} as follows:
\begin{definition}[CI-approximation~\cite{chanyeol_probabilistic}]
Given an RSTL formula $\Phi$, the CI-approximation of $\mathcal{P}(\Phi_{t} \mid x)$ is defined by: 
\begin{equation}
\begin{aligned}\label{eq:prob_sat}
\mathcal{P}(E^{i}_{t} \mid \mathbf{X}) &\equiv \mathcal{P}^{i}(\mathbf{X}, t)\\
\mathcal{P}(\neg \Phi_{t} \mid \mathbf{X}) &\equiv 1 - \mathcal{P}(\Phi_{t} \mid \mathbf{X})\\
\mathcal{P}(\bigwedge_{i} \Phi_{t}^{i} \mid \mathbf{X}) &\equiv \prod_{i} \mathcal{P}(\Phi_{t}^{i} \mid \mathbf{X})\\
% \mathcal{P}(\Phi_{t} \vee \Psi_{t} \mid x) &\equiv 1 - (1 - \mathcal{P}(\Phi_{t} \mid x) )( 1- \mathcal{P}(\Psi_{t} \mid x) )\\
% \mathcal{P}( (\mathcal{F}_{[t_{1}, t_{2}]}\Phi)_{t} \mid x) &\equiv 1 - \prod_{\tau \in t+[t_{1}, t_{2}]} (1 - \mathcal{P}(\Phi_{\tau} \mid x)) \\
\mathcal{P}( (\mathcal{G}_{[t_{1}, t_{2}]}\Phi)_{t} \mid \mathbf{X}) &\equiv \prod_{\tau \in t+[t_{1}, t_{2}]}\mathcal{P}(\Phi_{\tau} \mid \mathbf{X})\\
% \mathcal{P}(\Phi \mathcal{U}_{[t_{1}, t_{2}]} \Psi \mid \mathbf{x}_{1:T}, t) &\equiv 1 - \\
% \prod_{\tau_{1} \in t+[t_{1}, t_{2}]}\big(1 - (1 - \mathcal{P}(\Psi_{t} \mid \mathbf{x}_{1:T})) &\times ( 1- \prod_{\tau_{2} \in [t_{1}, t_{1} + \tau]} \mathcal{P}(\Psi_{\tau_{2}} \mid \mathbf{x}_{1:T}) )\big)
\end{aligned}
\end{equation}
% \mathcal{P}(\phi_{1} \mathcal{U} \phi_{2}, x, t) &\equiv 1 - \prod_{\tau_{1} \in t + [t_{1}, t_{2}]} ( 1 - \textbf{Sat}(x, \tau_{1}, \hat{\mathcal{E}}, \Psi) 
%     \vee &\bigwedge_{\tau_{2} \in [t_{1}, t_{1} + \tau_{1}]} \textbf{Sat}(x, \tau_{2}, \hat{\mathcal{E}}, \Phi),
\end{definition}
%%%%%%%%%%%%%%%%%%%%%%%%%%%%%%%%%%%%%%%%%%%%%%%%%%
\subsection{Log-odds Transform}

The output range for CI-approximation of~$\mathcal{P}$~\eqref{eq:prob_sat} is~$[0, 1]$ since it computes probability. 
This can lead to numerical instability, and gradient ascent often leads to poor convergence.
A natural re-parameterisation for Bernoulli random variables is the \emph{log-odds}:
\begin{equation}\label{eq:logit_sat:pred}
    \mathcal{L}(A) = \log \frac{ \mathcal{P}(A) }{ \mathcal{P}(\neg A) }
    ,
\end{equation}

It can be shown with some algebraic manipulations that re-writing the CI rule for pairwise disjunction~$\vee$ in~\eqref{eq:prob_sat} in terms of log-odds leads to:
% With some algebraic manipulation on~\eqref{eq:prob_sat}, a pair-wise disjunction in terms of log-odds is given by:
\begin{equation}\begin{aligned}
    \mathcal{L}(A \vee B) &= \log \frac{\mathcal{P}(A)\mathcal{P}(B) + \mathcal{P}(\neg A)\mathcal{P}(B) + \mathcal{P}(A)\mathcal{P}(\neg B)}{\mathcal{P}(\neg A)\mathcal{P}(\neg B)}  \\
                           &= \mathfrak{lse}(\mathcal{L}(A), \mathcal{L}(B), \mathcal{L}(A) + \mathcal{L}(B)) \label{eq:logit_sat:or},     
    %\log \frac{ \mathcal{P}(\Psi_{t} \mid x)\mathcal{P}(\mathcal{P}(\Psi_{t} \mid x)\mathcal{P}(\Phi_{t} \mid x) \Phi_{t} \mid x) + \mathcal{P}(\neg \Psi_{t} \mid x)\mathcal{P}(\Phi_{t} \mid x) }{ \mathcal{P}(\neg \Psi_{t} \mid x)\mathcal{P}(\neg \Phi_{t} \mid x) } \\
\end{aligned} \end{equation}
where $A$ and $B$ are independent Bernoulli random variables, and $\mathfrak{lse}$ is the \emph{log-sum-exp} function:
\begin{equation}
    \mathfrak{lse}(\mathcal{L}_{1}, ..., \mathcal{L}_{N})  = \log \sum_{i} \exp{\mathcal{L}_{i}}
    .
\end{equation}
A series of disjunction operations~\eqref{eq:logit_sat:or} is then:
\begin{equation}\label{eq:sum_over_all_subsets} \begin{aligned}
    \mathcal{L} \left( \bigvee_{i \in I} A_{i} \right) &= \log \sum_{J \in 2^{I}} \exp \sum_{j \in J} \mathcal{L}(A_{j})
    ,
\end{aligned} \end{equation} 
where $2^{I}$ denotes the power set of $I$. 

Computing log-sum-exp over sum of all subsets is clearly cumbersome.
We avoid such computation by using the relationship between elementary symmetric polynomials and monic polynomials. 
Observe that the summations over $j \in J$ can be taken out of the exponential as products.
Then, we have all elementary symmetric polynomials over $A_{i}$ less 1.
We thus arrive at a more compact expression:
% As we are summing all elementary symmetric polynomials except 1, we can aggregate the terms into a monic polynomial and arrive at:
\begin{equation}\label{eq:monic_polynomial}
    \mathcal{L} \left( \bigvee_{i} A_{i} \right) = \log \left( \prod_{i} \big(1 + \exp\mathcal{L}(A_{i})\big) - 1 \right). 
\end{equation}

% Combining these, we can evaluate the rest of the operators:
% \begin{align}
%     \mathbb{L}( \neg \phi, x, t) &= - \mathbb{L}( \phi, x, t)\label{eq:logit:not}\\ 
%     \mathbb{L}( \phi \vee \psi, x, t) &= \text{LSE}(\mathbb{L}( \phi, x, t), \mathbb{L}( \psi, x, t), \mathbb{L}( \phi, x, t) + \mathbb{L}( \psi, x, t) )\label{eq:logit:and} \\
%     \mathbb{L}( F_{[t_{1}, t_{2}]}\phi, x, t) &= \log \sum_{I \subset [t_{1}, t_{2}]} \exp(\sum_{t' \in I}\mathbb{L}( \phi, x, t') )  \\
% \end{align}
Now, the CI computation rules in the log-odds domain are given as follows:
\begin{definition}[CI-approximate log-odds]
Given an RSTL formula $\Phi$, the CI-approximation of log-odds of satisfaction $\mathcal{L}(\Phi_{t} \mid \mathbf{X})$ is calculated by: 
\begin{equation}\label{eq:logits_sat}
\begin{aligned}
\mathcal{L}(E_{t}^{i} \mid \mathbf{X}) &\equiv \log \mathcal{P}^{i}(\mathbf{X}, t) -  \log (1 - \mathcal{P}^{i}(\mathbf{X}, t)) \\
\mathcal{L}(\neg \Phi_{t} \mid \mathbf{X}) &\equiv - \mathcal{L}(\Phi_{t} \mid \mathbf{X})\\
\mathcal{L} \left( \bigvee_{i}\Phi_{t}^{i} \mid \mathbf{X} \right) &\equiv \log \left( \prod_{i} (1 + \exp\mathcal{L}(\Phi_{t}^{i} \mid \mathbf{X})) - 1 ) \right) \\% \mathcal{L}(\Phi_{t} \wedge \Psi_{t} \mid x) &\equiv -\mathfrak{lse}(-\mathcal{L}( \Phi_{t} \mid x), -\mathcal{L}(\Psi_{t} \mid x), -\mathcal{L}( \Phi_{t} \mid x) - \mathcal{L}(\Psi_{t} \mid x))\\
\mathcal{L}( (\mathcal{F}_{[t_{1}, t_{2}]}\Phi)_{t} \mid \mathbf{X}) &\equiv \log \Big( \prod_{\mathclap{\tau \in t + [t_{1}, t_{2}]}} \big(1 + \exp\mathcal{L}(\Phi_{\tau} \mid \mathbf{X}\big) - 1 \Big). \\
% \mathcal{L}( (\mathcal{G}_{I}\Phi)_{t} \mid x) &\equiv -\log( \prod_{i} (1 + \exp-\mathcal{L}(\phi_{i}, x, t)) - 1 ))
% \mathcal{L}(\Phi_{t} \mathcal{U} \Psi_{t} \mid x) &\equiv  -\log( \prod_{i} (1 + \exp-\mathcal{L}(\Phi^{i}\mid \mathbf{x}_{1:T})) - 1 ))
% \mathcal{L}( (\Phi\mathcal{U}\Psi)_{t} \mid \mathbf{X}) &\equiv \mathcal{L}( \bigvee_{\tau \in }  \Phi_{t} \mathcal{U} \Psi_{t} \mid \mathbf{X}) 
\end{aligned}
\end{equation}
\end{definition}
% \begin{remark}[Similarities to Deterministic STL]
% \hfill

% \textit{Similarities to deterministic STL:} 
It is interesting to note that the proposed computation rules for probability of satisfaction exhibits strong similarities to existing work on deterministic STL synthesis and model checking~\cite{donze_stl,lindermann,smooth_cumulative,belta_amgm}. In the log-odds domain, certain satisfaction (i.e. probability of 1) translates to $\infty$, certain dissatisfaction is $-\infty$, and absolute uncertainty (i.e. probability of 0.5) is $0$, which are the behaviours of spatial robustness measure for deterministic STL introduced in~\cite{donze_stl}. 
Further, the log-sum-exp function has been used in deterministic STL synthesis~\cite{lindermann,smooth_cumulative} as a smooth approximation of the maximum function. 
Finally, A similar expression to~\eqref{eq:monic_polynomial} was presented in~\cite{belta_amgm} as an alternative robustness measure for deterministic STL. The authors report encouragement of repeated satisfaction, which is consistent with the probability of disjunction increasing with increasing probability of the disjuncts.
    % aforementioned property of disjunction in the probabilistic setting. 
% \end{enumerate}
% \end{remark}

Given that the approaches that do not encourage repeated satisfaction~\cite{lindermann,smooth_cumulative} still report acceptable results, we consider the following approximation for disjunction:
\begin{equation}
\begin{aligned}
        \mathcal{L}(A \vee B) &\approx \mathfrak{lse}(\mathcal{L}(A), \mathcal{L}(B))) \\
        &= \log{ \frac{\mathcal{P}(A) \mathcal{P}(\neg B) + \mathcal{P}(\neg A) \mathcal{P}( B) }{\mathcal{P}(\neg A) \mathcal{P}(\neg B)} }
  .
    % &= \mathfrak{lse}(\mathcal{L}(\Phi_{t}\mid x), \mathcal{L}(\Psi_{t}\mid x), \mathcal{L}(\Phi_{t}\mid x) + \mathcal{L}(\Psi_{t}\mid x)) \\
\end{aligned}
\end{equation}
This ignores repeated satisfaction by omitting the $\mathcal{P}(A)\mathcal{P}(B)$ term from the numerator of~\eqref{eq:logit_sat:or}.
Meanwhile, there is a potential numerical benefit that log-sum-exp can be computed numerically stably with the so-called \emph{log-sum-exp trick}, while the product in~\eqref{eq:monic_polynomial} may underflow. 
We thus define the \emph{mutually exclusive}~(ME) approximation as follows. 
% \newpage
\begin{definition}[ME approximation]
Given an RSTL formula $\Phi$, the ME approximation of $\mathcal{L}(\Phi_{t} \mid \mathbf{X})$ is calculated by: 
\begin{equation}\label{eq:logits_approx_sat}
\begin{aligned}
\mathcal{L}(E_{t}^{i} \mid \mathbf{X}) &\equiv \log \mathcal{P}^{i}(\mathbf{X}, t) -  \log (1 - \mathcal{P}^{i}(\mathbf{X}, t)) \\
\mathcal{L}(\neg \Phi_{t} \mid \mathbf{X}) &\equiv - \mathcal{L}(\Phi_{t} \mid \mathbf{X})\\
\mathcal{L}(\Phi_{t} \vee \Psi_{t} \mid \mathbf{X}) &\equiv \mathfrak{lse}(\mathcal{L}( \Phi_{t} \mid x), \mathcal{L}(\Psi_{t} \mid \mathbf{X}))\\
% \mathcal{L}(\Phi_{t} \wedge \Psi_{t} \mid x) &\equiv -\mathfrak{lse}(-\mathcal{L}( \Phi_{t} \mid x), -\mathcal{L}(\Psi_{t} \mid x))\\
% \mathcal{L}(\Phi_{t} \mathcal{U} \Psi_{t} \mid x) &\equiv  \log( \prod_{i} (1 + \exp\mathcal{L}(\phi_{i}, x, t)) - 1 ))\\
\mathcal{L}( (\mathcal{F}_{I}\Phi)_{t} \mid \mathbf{X}) &\equiv \log \sum_{\tau \in t + I} \exp\mathcal{L}(\Phi_{\tau} \mid \mathbf{X}).
% \mathcal{L}( (\mathcal{G}_{I}\Phi)_{t} \mid x) &\equiv -\log \sum_{\tau \in t + I} \exp-\mathcal{L}(\Phi_{\tau} \mid x)
\end{aligned}
\end{equation}
\end{definition}
\begin{figure}[t]
    \centering
    \subfloat[$\mathcal{F}(A) \wedge \mathcal{F}(B)$\label{fig:soundness:FaFb}]{\includegraphics[width=0.45\columnwidth]{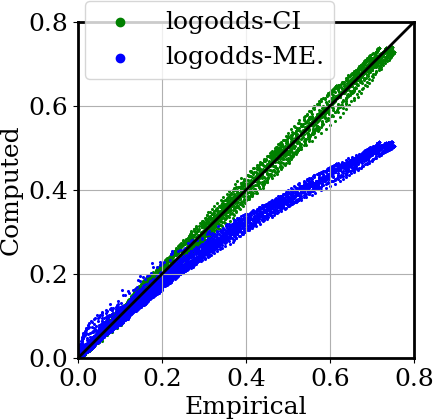}}
    \subfloat[$\mathcal{F}(A \wedge \mathcal{F}(B))$\label{fig:soundness:sequencing}]{\includegraphics[width=0.45\columnwidth]{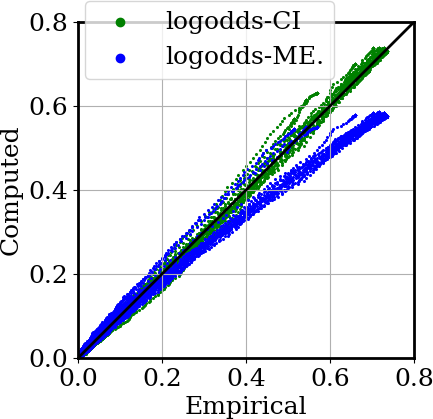}}
    % \subfloat[]{\includegraphics[width=0.23\textwidth]{Figures/soundness/soundness_FaFb.png}}\label{fig:convergence:notaUbandFa} 
    \caption{Comparison of the CI~(green) and ME~(blue) approximation results against MC estimates~(`Empirical'). CI is exact for $\mathcal{F}(A) \wedge \mathcal{F}(B)$, while ME underestimates. For $\mathcal{F}(A \wedge \mathcal{F}(B))$, the error increases, but not significantly. Naive method's result showed numerically insignificant difference to CI, and is omitted.}
    \label{fig:soundness}
\end{figure}
%%%%%%%%%%%%%%%%%%%%%%%%%%%%%%%%%%%%%%%%%%%%%%%%%%
\subsection{Synthesis with Gradient-Ascent}
With the probability or log-odds of satisfaction computed, we synthesise a MAP control sequence~$\mathbf{U}^*$ that maximises the posterior probability. 
We use Jensen's inequality to bound the log of posterior probability~\eqref{eq:stl_map}: 
\begin{equation}
\begin{aligned}
    \log \mathcal{P}( \mathbf{U} \mid \Phi_{t} ) \geq &\underset{\mathbf{X}  \sim \mathcal{P}(\mathbf{X}  \mid \mathbf{U} )}{\mathbb{E}}[\log \mathcal{P}( \Phi_{t} \mid \mathbf{X}) ) ] 
                                                            +\log \mathcal{P}(\mathbf{U} ).
\end{aligned}
\end{equation}

Subsequently, we maximize the lower bound:
\begin{equation}\label{eq:expected_log_prior}
    \mathbf{U}^{*} = \argmax_{\mathbf{U}} \underset{\mathbf{X} \sim \mathcal{P}(\mathbf{X} \mid \mathbf{U})}{\mathbb{E}} \left[ \log \mathcal{P}( \Phi_{t} \mid \mathbf{U}) \right]  + \log \mathcal{P}(\mathbf{U}).
\end{equation}
\begin{figure*} [t]
    \centering
    \subfloat[$N_{s}=1$]{\includegraphics[width=0.32\textwidth]{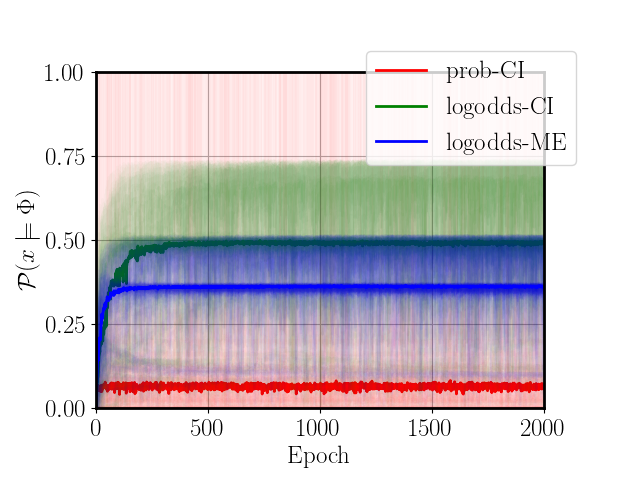}}\label{fig:convergence:ns-1}
    % \subfloat[$N_{s}=10$]{\includegraphics[width=0.23\textwidth]{Figures/convergence/convergence-Ns-10-upto-2000.png}}\label{fig:convergence:ns-10} \\
    \subfloat[$N_{s}=50$]{\includegraphics[width=0.32\textwidth]{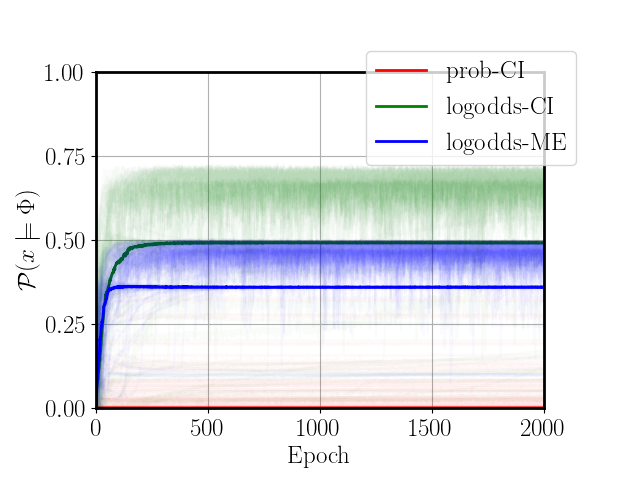}}\label{fig:convergence:ns-50}
    \subfloat[$N_{s}=100$]{\includegraphics[width=0.32\textwidth]{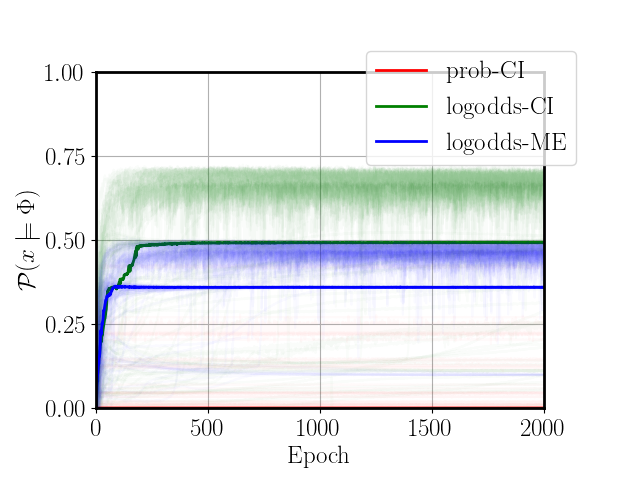}}\label{fig:convergence:ns-100}
    \caption{Comparison of convergence with  number of trajectory samples. Solid lines are medians.}
    \label{fig:convergence}
\end{figure*}

The maximisation is done by gradient ascent on~\eqref{eq:expected_log_prior}. 
Because the expectation in~\eqref{eq:expected_log_prior} is intractable, we replace it with an empirical mean over a $N_{s}$ number of trajectory samples, so that the $i$-th gradient ascent step is: 
\begin{equation}\label{eq:gradient_update}
    \hat{\mathbf{U}}^{i+1} = \hat{\mathbf{U}}^{i} +  \frac{1}{N_{s}} \sum_{j} \frac{\partial}{\partial \hat{\mathbf{U}}^{i} } [ \log \mathcal{P}( \Phi_{t} \mid \mathbf{X}_{1:T}^{j}(\hat{\mathbf{U}}^{i} )) + \log \mathcal{P}(\hat{\mathbf{U}}^{i} ) ]
    ,
\end{equation}
where $\hat{\mathbf{U}}^{i+1}=\begin{bmatrix}\hat{\mathbf{u}}_{1}^{i}...\hat{\mathbf{u}}_{T}^{i}\end{bmatrix}$.
Each trajectory sample $\mathbf{X}^{j}(\hat{\mathbf{U}}^{i})$ is obtained by propagating the dynamic model~\eqref{eq:dynamic_model} forward in time including actuation uncertainty.
Note that, as long as the predicates' distributions and the dynamic model are differentiable, so is~\eqref{eq:gradient_update}.
For both CI and ME approximations, analytical gradients can be computed easily using autograd frameworks such as Tensorflow~\cite{tensorflow} or PyTorch~\cite{pytorch,karen_leung}.
Therefore, we do not present the expressions here. 

%%%%%%%%%%%%%%%%%%%%%%%%%%%%%%%%%%%%%%%%%%%%%%%%%%
%%%%%%%%%%%%%%%%%%%%%%%%%%%%%%%%%%%%%%%%%%%%%%%%%%
% \newpage
\section{Empirical Analysis}
% We discuss various algorithmic aspects of the proposed framework. 

We evaluate the practical performance characteristics of the proposed method. 
We first demonstrate that the proposed CI~\eqref{eq:logits_sat} and ME~\eqref{eq:logits_approx_sat} methods reasonably approximate the ground truth~\eqref{eq:prob_sat_monte_carlo}. 
We then examine the convergence characteristics of the gradient ascent~\eqref{eq:gradient_update} solution, and demonstrate the computational benefits of GPU acceleration. 
% We then show that the framework is GPU-compatible that it finds a better solution by concurrently running many initial conditions. 

\begin{figure} [t]
    \centering
    \includegraphics[width=0.6\columnwidth]{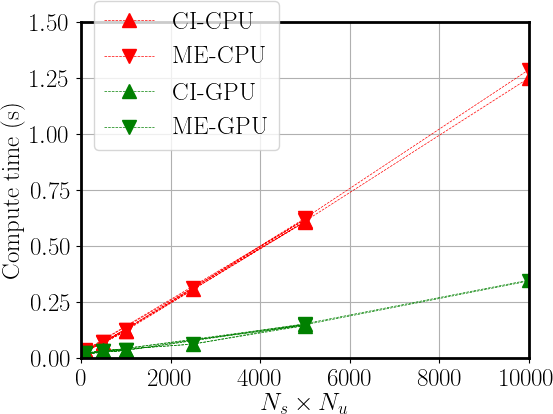} 
    \caption{Comparison of average computation time per gradient ascent step between combinations of CPU~(red) and GPU~(green) with CI~(upward triangle and ME~(downward triangle). GPU shows 4-fold improvement in scalability. Variance was in the order of $10^{-4}$ for all configurations.}
    \label{fig:computation}
\end{figure}

%%%%%%%%%%%%%%%%%%%%%%%%%%%%%%%%%%%%%%%%%%%%%%%%%%
\subsection{Quality of Approximation}

% We first postulate a theoretical hypothesis, and test these hypotheses empirically. 
In this section, we evaluate whether the CI and ME approximations compute the probability of satisfaction accurately. 
Because the CI computation rule assumes independence amongst operands, we expect it to be exact if 1) all predicates are conditionally independent; 2) each predicate is independent across time and space; and 3) only one operator uses each predicate.
For example, assuming 1) and 2) hold, we expect the CI rule to be exact on $\mathcal{F}A \wedge \mathcal{F}B$, but not $\mathcal{F}(A \wedge \mathcal{F}B)$, because the disjuncts of the outer $\mathcal{F}$ operator are not independent. 
The ME computation rules~\eqref{eq:logits_approx_sat} will not be exact in any case. 
% This is because the CI method assumes independence between operands, which is true if the operators act on disjoint sets of predicates and the predicate independence assumption holds.

We validate these hypotheses by comparing against a 1000-sample Monte Carlo~(MC) approximation of RSTL probability of satisfaction~\eqref{eq:prob_sat_monte_carlo}.
We used the trajectories from the first 2000 gradient ascent steps generated from the target search scenario (Fig.~\ref{fig:two_targets}).  
For simplicity, we evaluated each predicates' marginal probability independently before sampling, so that the first two conditions of exactness hold. 

In Fig.~\ref{fig:soundness:FaFb}, CI (green) and ME (blue) results are compared against the MC estimate for $\mathcal{F}A \wedge \mathcal{F}B$.  
It can be seen that the CI method matches the MC result as expected, while ME consistently underestimates.
This is expected, because ME does not account for multiple satisfaction. 

Figure~\ref{fig:soundness:sequencing} shows comparison for $\mathcal{F}(A \wedge \mathcal{F}B)$.
% Although we expect neither CI nor ME to be exact,  albeit with increased error than the previous case.
As $\mathcal{F}B$ is double-counted by the outer $\mathcal{F}$, CI and ME tend to over-estimate, but not by much.
ME continues to underestimate, and CI matches the MC closely,
showing that CI and ME are reasonable approximations for practical applications. 

% The result is shown in Fig.~\ref{fig:soundness}. 
% It can be seen that the "full" computation rule is in close agreement with the Monte Carlo result, with no apparent bias.
% This is expected, because all the operands of the formula $\mathcal{F}(a) \vee \mathcal{F}(b)$ are independent, except that the predicates are related over time due to the continuity of targets' trajectory. 
% The variances are attributed to the Monte Carlo nature of the `empirical ground truth' we compare to. 
% Meanwhile, the "approximate" computation rule consistently underestimates the probability. 
% This is expected, as the approximate rule is obtained by ignoring multiple satisfactions. 
% We note, however, that the approximate computation rule remains to monotonically increasing with ground truth probability. 
% This implies that, in the context of synthesis, using the approximate computation rule should yield the same trajectory, because a monotone transform of a scalar function does not alter the maximum. 
% Admittedly, this need not hold true for all STL formulae, but we hypothesise it does for a certain useful class of formulae. 

%%%%%%%%%%%%%%%%%%%%%%%%%%%%%%%%%%%%%%%%%%%%%%%%%%
\subsection{Convergence}\label{sec:convergence}

Gradient-based methods cannot guarantee globally optimal solutions unless the objective is convex.  
We analyse the convergence characteristics of the proposed computation rules in the target search scenario~(Fig.~\ref{fig:two_targets}). 
We randomly generated 100 initial conditions from the control prior $\mathcal{P}(\mathbf{U})$, and ran the gradient ascent step for 2000 iterations, with $N_{s} = 1, 50, 100$ number of trajectory samples. 

Figure~\ref{fig:convergence} shows the probability of satisfaction over gradient ascent steps for naive CI~(\eqref{eq:prob_sat}, red), log-odds CI~(\eqref{eq:logits_sat}, green) and log-odds ME~(\eqref{eq:logits_approx_sat}, blue) methods with varying number of trajectory samples $N_{s}$. 
It can be seen that while log-odds CI and ME methods find global and local optima, the naive CI method does not find any.
The naive CI method's failure is attributed to the computation rules~\eqref{eq:prob_sat} being bounded to $[0, 1]$ range, which causes numerical errors to build up. 
Note that log-odds ME reports lower probability of satisfaction due to its underestimation property, but we found that the resulting trajectories were still similar.

With increasing $N_{s}$, the variance in probability of satisfaction is decreased. 
In practice, this means the generated plan will more reliably account for state uncertainty, which is useful for, e.g., the collision avoidance scenario in Fig.~\ref{fig:indoor}. 
% The result shows that our method outperformed the rest in all ranges of~$N_s$.
% For the naive method (prob-CI) where the output ranges from 0 to 1, the gradient-ascent scheme suffered from numerical instability that it performed dramatically worse compared to other methods. 
% It is clear that the increase in the number of state samples~$N_s$ reduces the variance in estimation. It implies that increasing the number of samples is necessary to improve the convergence.
% The first notable observation is that the naive method does not perform very well. 
% Remarkably, although the log-odds CI and naive CI methods predict the same probability, the log-odds method  
% This is expected, because performing gradient updates with bounded parameters often suffer from numerical issues. 
% We also note that the CI shows slightly higher variance than the ME method at the verge of convergence to optima. 
% This is most likely due to numerical underflows in the monic polynomial computation. 
% Further, note that the local optimum is dominant, and the median being at the local optimum.
% We also note that the variance of estimated probability of satisfaction improves with more number of state samples.
% These observations show the importance of using multiple initial conditions and an adequate number of trajectory samples. 
%  imply that for practical applications, it is import to
% %%%%%%%%%%%%%%%%%%%%%%%%%%%%%%%%%%%%%%%%%%%%%%%%%%
\subsection{Computation Time}

The results in Sec.~\ref{sec:convergence} illustrate that it is important to use multiple initial conditions and more trajectory samples to ensure reliable operation. 
However, this would inevitably increase the computation time as well. 

% gradient-ascent scheme, in general, relies on the number of initial conditions in order to avoid local optimum. In our problem, we observe that the estimated probability of satisfaction improves with more number of state samples. 
% These aspects imply that the convergence rate will increase with the number of both initial conditions and state samples. 
GPU acceleration is a prominent means to circumvent the issue of computation time, but not all algorithms benefit from GPU acceleration. 
To determine if our proposed methods benefit from GPU acceleration, we compare the computation time per gradient ascent step of our Tensorflow~\cite{tensorflow} implementation between GPU and CPU.
We used all combinations between $N_{s} = 1, 10, 50, 100$ and $N_{u} = 1, 10, 50, 100$, and computed the mean over 1000 gradient ascent steps.
We used a desktop with CPU~(Intel i5-9500) and a GPU~(NVIDIA RTX2060) to conduct the experiment.

Figure~\ref{fig:computation} shows the computation time with varying number of initial conditions~$N_u$ and the number of state samples~$N_s$. 
We found that the total number of samples $N_{u} \times N_{s}$ explains all changes. 
The result shows that the computation time with GPU is significantly lower than that with CPU, and that using a GPU leads to $4$-fold improvement in scalability.
This demonstrates that the proposed gradient ascent method benefits from GPU acceleration. 
% The computation time scales linearly
% The performance could be improved by running the framework with more GPU processors, the framework is tractable in those variables that affect the convergence rate.
% As the convergence results show the importance of multiple initial conditions and trajectory samples, it is imperative to evaluate how the computation time scales with these parameters. 
% In particular, we evaluate how computation time scales on a CPU~(Intel i5-9500) and a GPU~(NVIDIA RTX2060). 
% We measured the computation time of initial conditions across 1000 iterations. 
%%%%%%%%%%%%%%%%%%%%%%%%%%%%%%%%%%%%%%%%%%%%%%%%%%
%%%%%%%%%%%%%%%%%%%%%%%%%%%%%%%%%%%%%%%%%%%%%%%%%%
\section{Case Studies}

We demonstrate two example cases that illustrate the benefit of using RSTL for task specification in uncertain environments.
The proposed gradient ascent method was implemented in Tensorflow~\cite{tensorflow}. 
For all examples, we consider a robot described by a bicycle dynamic model:
\begin{equation} \begin{aligned}
    \dot{\mathbf{x}}_t = \begin{bmatrix}
        \dot{x}_{t} \\
        \dot{y}_{t} \\
        \dot{\theta}_{t}
    \end{bmatrix} = 
    \begin{bmatrix}
        V_{t} \cos \theta_{t} \\ 
        V_{t} \sin \theta_{t} \\
        \omega_{t} + \epsilon_{t}
    \end{bmatrix}
    % \dot{x}_{t} &= V_{t} \cos \theta_{t} \\ 
    % \dot{y}_{t} &= V_{t} \sin \theta_{t} \\
    % \dot{\theta}_{t} &= u_{t} + \epsilon_{t},
    ,
\end{aligned} \end{equation}
where $\epsilon_{t} \sim \mathcal{N}(0, \sigma_{u})$ is white Gaussian noise.
The control inputs are $\mathbf{u}_{t} = \begin{bmatrix}V_{t} & \omega_{t} \end{bmatrix}^{\top}$. 
% The objective is to find a discrete sequence of control~$\mathbf{u}^* = u_0 \cdots$ where~$u_t \in \mathbb{R}$ is found in continuous action space using the gradient-ascent scheme.

%%%%%%%%%%%%%%%%%%%%%%%%%%%%%%%%%%%%%%%%%%%%%%%%%%
\subsection{2D Target Search}\label{sec:application:evaders}

\begin{figure}[t]
\centering
    % \subfloat[$t=15$]{\includegraphics[width=0.45\columnwidth]{Figures/two_targets/two-targets-good-t-15-crop.png}}\label{fig:} 
    % \subfloat[$t=15$]{\includegraphics[width=0.45\columnwidth]{Figures/two_targets/two-targets-bad-t-15-crop.png}}\label{fig:} \\

    \subfloat[$t=15$\label{fig:two_targets_good_25}]{\includegraphics[width=0.45\columnwidth]{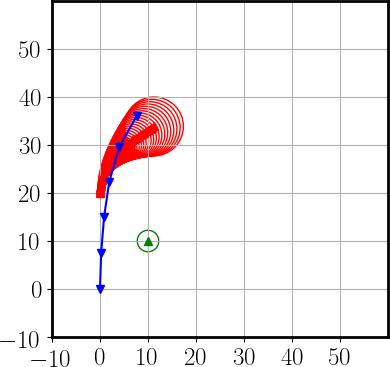}} 
    \subfloat[$t=10$\label{fig:two_targets_bad_25}]{\includegraphics[width=0.45\columnwidth]{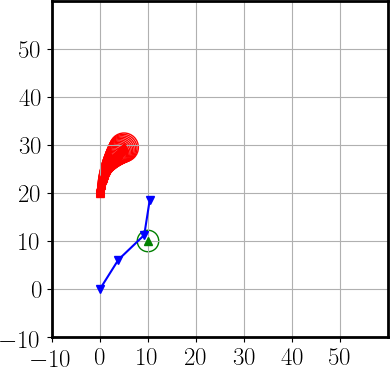}} \\

    % \subfloat[$t=35$]{\includegraphics[width=0.45\columnwidth]{Figures/two_targets/two-targets-good-t-35-crop.png}}\label{fig:two_targets_good_35} 
    % \subfloat[$t=35$]{\includegraphics[width=0.45\columnwidth]{Figures/two_targets/two-targets-bad-t-35-crop.png}}\label{fig:two_targets_bad_35} \\

    \subfloat[$t=45$\label{fig:two_targets_good_45}]{\includegraphics[width=0.45\columnwidth]{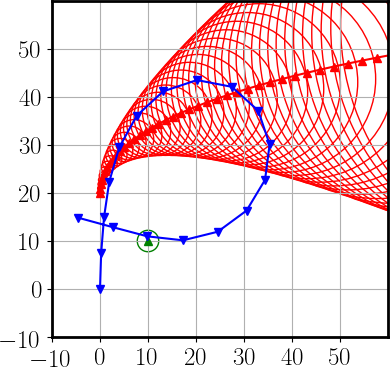}} 
    \subfloat[$t=45$\label{fig:two_targets_bad_45}]{\includegraphics[width=0.45\columnwidth]{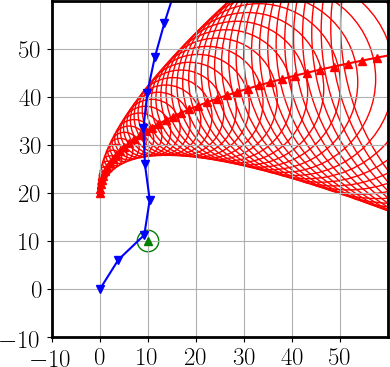}} \\
    \caption{A 2D target search scenario. The robot~(blue) is tasked with detecting both Tom~(Green) and Jerry~(Red). The global optimum with $\mathcal{P}(\Phi \mid \mathbf{x}) \approx 0.7$ ~(left column) is to detect Jerry before uncertainty grows. A local optimum with $\mathcal{P}(\Phi \mid \mathbf{x}) \approx 0.5$~(right column) prefers Tom, who is closer. Circles show 1-covariance bound. \label{fig:two_targets}}
\end{figure}

\begin{figure}[t]
    \centering
    % \subfloat[`Visit all targets with no collision' (without control noise, $\sigma_{u}=10^{-4}$)]{\includegraphics[width=0.48\columnwidth]{Figures/indoor/targets_no_noise.png}}\label{fig:indoor:no_noise} 
    % \subfloat[`Visit all targets while avoiding collision' (with control noise $\sigma_{u}=0.1$)]{\includegraphics[width=0.48\columnwidth]{Figures/indoor/targets_noise.png}}\label{fig:indoor:noise} \\
    \subfloat[$\sigma_{u}=10^{-4}$\label{fig:indoor:complex_no_noise}]{\includegraphics[width=0.48\columnwidth]{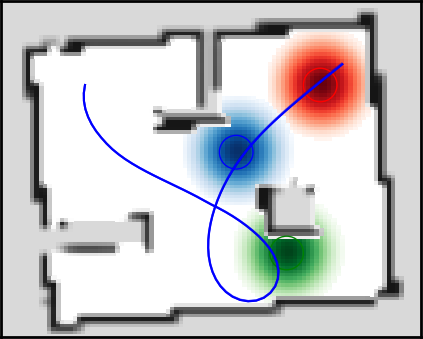}} 
    \subfloat[$\sigma_{u}=0.1$\label{fig:indoor:complex_noise}]{\includegraphics[width=0.48\columnwidth]{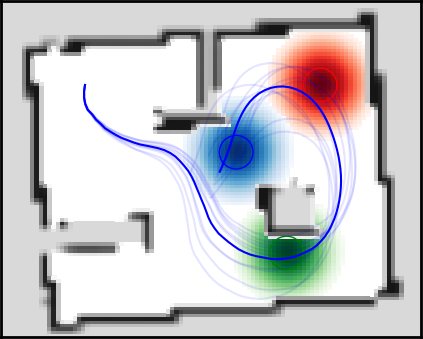}}     
    \caption{Results for complex indoor mission. The robot's~(blue) task is a conjunction of `never visit Rob~(red) or Bob~(blue) before visiting sanitising station~(green), and avoid obstacles~(grey colormap)', and 'visit Rob~(red) and Bob~(blue)'. With higher actuation uncertainty~$\sigma_{u}$, the trajectory becomes further from the walls. Solid lines are the robot's nominal trajectory in the absence of noise. Transparent blue lines are the noised samples used during synthesis. The start location is top-left corner.\label{fig:indoor}}
    
\end{figure}

We consider a 2D target search scenario, where a robot is tasked with detecting possibly moving targets in the environment: \emph{Tom} and \emph{Jerry}. Jerry, as usual, is moving with increasing uncertainty, while Tom is stationary with high certainty. Figure~\ref{fig:two_targets} depicts an example where the mean paths for Tom and Jerry are shown in green and red. The growing uncertainty over time is shown around the mean. Note that since Tom (in green) is known to be stationary, its uncertainty does not grow over time. The robot starts at~$[0, 0]^\top$.

The task of finding Tom and Jerry can be expressed using RSTL as $\Phi = \mathcal{F}(D^{\text{Tom}}) \wedge \mathcal{F}(D^{\text{Jerry}})$~(i.e., `eventually detect Tom and eventually detect Jerry').

We model the events $D^{\text{Tom}}$ and $D^{\text{Jerry}}$ as follows. 
If the location is known, the robot detects Tom and Jerry with likelihood modelled by:
\begin{equation}\label{eq:detection_likelihood}
    \mathcal{P}(D_{t} \mid \mathbf{x}_{t}, \mathbf{z}_{t}) = P_{D} \exp \left( \frac{|| \mathbf{x}_{t} - \mathbf{z}_{t} ||^{2}} {2 r_{D}^{2}} \right)
    , 
\end{equation}
where $\mathbf{z}_{t}$ is the location of the target, $r_{D}$ is the radius of detection, and $P_{D}$ controls the peak. 

Tom and Jerry are modelled by a constant acceleration model, which is a linear Gaussian system. 
The mean $\bar{\mathbf{z}}^{a,b}_{t}$ and $\Sigma^{a,b}_{t}$ are propagated given the robots' belief at $t=0$ using the standard prediction equations.
The marginal probability of detection accounting for their uncertainty is given by:
\begin{equation}\label{eq:pred:detection}
\begin{aligned}
    \mathcal{P}(D_{t} \mid \mathbf{x}_{t}) &= \int \exp \left( \frac{|| \mathbf{x}_{t} - \bar{\mathbf{z}}_{t} ||^{2}} {2 r_{d}^{2}} \right) \mathcal{N}(\bar{\mathbf{z}}_{t}, \Sigma^{z}_{t}) d\mathbf{z}_{t} \\
                                 &= \sqrt{2\pi^{N}} r_{d}^{N} \mathcal{N}(\bar{\mathbf{z}}_{t}, \Sigma^{z}_{t} + r_{d}^{2} I), 
\end{aligned}    
\end{equation}
where $\mathcal{N}(\bar{\mathbf{z}}_{t}, \Sigma^{z}_{t})$ is the multivariate Gaussian PDF. 

% Since the gradient-ascent scheme finds the empirical mean over a fixed number of trials, the resulting solution is subject to stochasticity. 
Figure~\ref{fig:two_targets} shows global and local optima found using the log-odds CI computation rule~\eqref{eq:logits_sat}.
It can be seen that the global optimum~(Figs.~\ref{fig:two_targets_good_25} and~\ref{fig:two_targets_good_45}) is to detect Jerry first at $t=15$~(Fig.~\ref{fig:two_targets_good_25}), and to return to Tom at $t=45$~(Fig.~\ref{fig:two_targets_good_45}), while the local optimum is to detect Tom first at $t=10$~(Fig.~\ref{fig:two_targets_bad_25}) and then Jerry later. 
This is because the uncertainty of Jerry grows unlike Tom, and the optimal trajectory should detect Jerry first before its uncertainty grows. 
This demonstrates that RSTL naturally reasons over uncertainty. 
%%%%%%%%%%%%%%%%%%%%%%%%%%%%%%%%%%%%%%%%%%%%%%%%%%
\subsection{Complex Missions in an Indoor Environment}\label{sec:application:indoor}
% We consider complex missions in a realistic 2D indoor environment commonly used as a benchmark in perception research~\cite{lan_and_me,gpom,bhoram}.
% Figure~\ref{fig:indoor} shows the 2D indoor environment where the probabilistic likelihood of finding each of them is shown in red, blue and green colours; the darker colour represents higher likelihood. A robot path is shown with blue line.
Consider a nursing robot in an indoor environment, modelled as an occupancy grid $O$ such that $O_{i,j}$ is the probability of obstacle occupancy. 
Collision with an obstacle is modelled by an interpolation:
\begin{equation}
    \mathcal{P}(O \mid \mathbf{x}_{t}) = \sum_{i,j} K_{ij}(\mathbf{x}_{t})O_{i, j},
\end{equation}
where $K_{ij}(\mathbf{x})$ denotes the interpolant.
%\footnote{We used Tensorflow's \textit{batch\_interp\_regular\_nd\_grid}, which is a differentiable operation}. 

The robot cares for two patients, Rob and Bob. 
The doctor asks the robot to avoid obstacles, and to never visit any of the patients before visiting the sanitising station, which can be written as an RSTL formula:
\begin{equation}
    \Phi_{1} = (\neg( D^{Rob} \vee D^{Bob} ) \mathcal{U} D^{San} ) \wedge \mathcal{G}(\neg O),
\end{equation}
where $D^{Rob}$, $D^{Bob}$, and $D^{San}$ are distributed as per~\eqref{eq:pred:detection}.

Now, in addition to the previous command, the doctor asks the robot to visit the two patients:
\begin{equation}
    \Phi_{2} = \mathcal{F}(D^{Rob}) \wedge \mathcal{F}(D^{Bob}) \wedge \Phi_{1}. 
\end{equation}

% Collision avoidance is handled by an STL formula, $\Phi^{\text{col}} = \mathcal{G}(\neg C)$~(\textit{always do not collide}). 
We created an occupancy grid from a realistic dataset commonly used in perception research~\cite{Bhoram,lan_and_me}, and compared the results with low~($10^-4 rads^{-1}$) and high~($\sigma_{u} = 0.1 \text{rad}s^{-1}$) actuation uncertainty.
The results are shown in Fig.~\ref{fig:indoor}.
In both cases, the generated trajectory is correct, visiting the sanitising station first, and then the two patients.
Interestingly, the trajectory changes drastically when control noise increases. 
The path with small control noise in Fig.~\ref{fig:indoor:complex_no_noise} is aggressively close to the wall, whereas the path with control noise in Fig.~\ref{fig:indoor:complex_noise} is more conservative in that the robot keeps distance away from the wall by manoeuvring around the obstacle.
This demonstrates that the proposed probabilistic formulation enables risk-averse behaviour in STL synthesis, a crucial property for practical applications.  
% It can be seen that both tasks are faithfully executed regardless of increase in control noise.
% All targets are visited, and particularly in Figs.~\ref{fig:indoor:complex_no_noise}~and~\ref{fig:indoor:complex_no_noise}, red and blue targets are visited \emph{after} the green target. 
% It is interesting to note that the trajectories vary with increase in $\sigma_{e}$, even for the same task.
% This is because with increasing control noise, the trajectories in the low-noise cases risk violating the no-collision condition, as 
% We also note that only ME approximation succeeded. 
% The CI approximation method failed due to a numerical underflow caused by the collision avoidance condition.
% To this end, let us first introduce the following assumptions on the random events $E$: 
% \begin{assumption}[Conditional independence\label{asm:cond_ind}]
% % $\forall \pi, \pi' \in \Pi$ and $t, t' \in \mathbb{R}^{+}$:
% % \begin{enumerate}
% $\forall E^{i} \neq E^{j} \in \mathcal{E}$, $E^{i}_{t}$ and $E^{j}_{t}$ are conditionally independent given $x$ for all $t$.
% Further, for all $t \neq t'$, $E^{i}_{t}$ and $E^{j}_{t}$ 
% % \end{enumerate}
% % maybe talk about this being a maximum entropy 
% \end{assumption}
%%%%%%%%%%%%%%%%%%%%%%%%%%%%%%%%%%%%%%%%%%%%%%%%%%
\section{Conclusion}
We have presented a probabilistic inference perspective on STL synthesis based on RSTL and corresponding algorithms.
%In doing so, we presented RSTL as a tool for task specification and synthesis of intelligent behaviours against uncertainty that are relevant to practical robotics applications. 
% RSTL extends the possible class of environmental and actuation uncertainty, and 
Our method exhibits appealing computational and expressivity characteristics that suit practical robotics applications.
We anticipate that the inference formulation of STL synthesis presented in this paper will accelerate the development of explainable AI techniques through seamless integration between formal methods and machine learning techniques as did the optimal control-as-inference paradigm~\cite{kappen,levine}. Many avenues of future work arise from our results; one of the most exciting is integration with estimation methods~\cite{lee2019online,best2015bayesian,rhinehart2019precog} that would allow multi-robot systems to augment or replace explicit communication for behaviour coordination with trajectory predictions derived from specifications~\cite{disr,decmcts}.

\bibliography{references}

\end{document}